\pdfoutput=1
\documentclass[11pt]{article}

\usepackage[]{acl}

\usepackage{times}
\usepackage{latexsym}

\usepackage[T1]{fontenc}
\usepackage[utf8]{inputenc}

\usepackage{microtype}

\usepackage{color}
\usepackage{booktabs}
\usepackage{multirow}
\usepackage{graphicx}
\usepackage{array}           
\usepackage{longtable, tabu}  
\usepackage{colortbl} 
\usepackage{tabularx}
\usepackage{enumitem}
\usepackage{amssymb}
\usepackage{subfigure}
\usepackage[nameinlink]{cleveref}
\crefformat{section}{\S#2#1#3} % see manual of cleveref, section 8.2.1
\crefname{algorithm}{Alg.}{Algs.}
\crefformat{subsection}{\S#2#1#3}
\Crefname{equation}{Eq.}{Eqs.}
\Crefname{figure}{Fig.}{Figs.}
\definecolor{Lightgray}{RGB}{110,110,110}

\title{IndiVec: An Exploration of Leveraging Large Language Models for Media Bias Detection with Fine-Grained Bias Indicators}

\author{Luyang Lin$^{1,2}$, Lingzhi Wang$^{1,2}$\thanks{~~Lingzhi Wang is the corresponding author.}, Xiaoyan Zhao$^{1}$, Jing Li$^3$, Kam-Fai Wong$^{1,2}$\\
  $^1$The Chinese University of Hong Kong, Hong Kong, China\\
  $^2$MoE Key Laboratory of High Confidence Software Technologies, China\\
  $^3$Department of Computing, The Hong Kong Polytechnic University, Hong Kong, China\\
  \tt $^{1,2}$\{lylin, lzwang, xzhao, kfwong\}@se.cuhk.edu.hk \\
  \tt $^3$jing-amelia.li@polyu.edu.hk\\
}

\begin{document}
\maketitle
\begin{abstract}
This study focuses on media bias detection, crucial in today's era of influential social media platforms shaping individual attitudes and opinions. In contrast to prior work that primarily relies on training specific models tailored to particular datasets, resulting in limited adaptability and subpar performance on out-of-domain data, we introduce a general bias detection framework, IndiVec, built upon large language models. IndiVec begins by constructing a fine-grained media bias database, leveraging the robust instruction-following capabilities of large language models and vector database techniques. When confronted with new input for bias detection, our framework automatically selects the most relevant indicator from the vector database and employs majority voting to determine the input's bias label. IndiVec excels compared to previous methods due to its adaptability (demonstrating consistent performance across diverse datasets from various sources) and explainability (providing explicit top-k indicators to interpret bias predictions). Experimental results on four political bias datasets highlight IndiVec's significant superiority over baselines. Furthermore, additional experiments and analysis provide profound insights into the framework's effectiveness.

\end{abstract}

\section{Introduction}
The widespread expansion of digital media platforms has introduced an era characterized by unparalleled accessibility to news and information. In today's digital era, misinformation and disinformation frequently gains traction on social media, thereby exerting a significant influence on public perception and decision-making. Given the critical impact of media bias on shaping attitudes and opinions, there exists a pressing need for the development of effective tools designed for detecting bias in media content.

\definecolor{left-leaning}{RGB}{220,20,60} 
\definecolor{right-leaning}{RGB}{0,0,128}   
\definecolor{neutral}{RGB}{100,100,100}     

\begin{table}[t]
\setlength{\tabcolsep}{2.5mm}\small
\begin{center}
\resizebox{\linewidth}{!}{
\begin{tabular}{lp{1.8cm}cp{3.2cm}}
\toprule[1.0pt]

 & & \textbf{Number} &\multicolumn{1}{>{\centering\arraybackslash}m{32mm}}{\textbf{Example}}\\

\midrule[0.5pt]
\multirow{4}{*}{\textbf{Framing}}&   \cite{card2015media} & 15 & Economic, Health and safety, Cultural identity  \\
\cmidrule(lr){2-4}
% \midrule[0.5pt]
% Persona \cite{card2016analyzing} & 50 &  REFUGEE immigrant boy, IMMIGRANT man alien \\
% \midrule[0.5pt]
& \cite{liu2019detecting} & 7 & Gun control/regulation, Mental health\\
\midrule[0.5pt]
% Indicator & & \\
\rowcolor[rgb]{0.93,0.93,0.93} 
   \multicolumn{2}{l}{ \textbf{Indicator (Ours)}}  & >20k& \multicolumn{1}{>{\centering\arraybackslash}m{32mm}}{$\triangledown$ Example $\triangledown$}  \\
\multicolumn{4}{p{9cm}}{ \textit{Sources and Citations}: Nielsen viewer data, TechCrunch online viewership - \textcolor{neutral}{Neutral} }  \\
\multicolumn{4}{p{9cm}}{\textit{Coverage and Balance}: Focuses on Republican Party divisions and criticisms of Trump - \textcolor{left-leaning}{Left Leaning}}  \\
\multicolumn{4}{p{9cm}}{\textit{Tone and Language}: Uses positive language to describe the expungement process and its potential benefits - \textcolor{right-leaning}{Right Leaning} }  \\
% IndiVec (Ours) & >20,000 & Nielsen viewer data, TechCrunch online viewership \\

\bottomrule[1.0pt]

\end{tabular}
}
\end{center}
% \vskip -1em
\caption{\label{tab:intro} Comparison of Framing and Bias Indicator. 
}
% \vskip -1em
\end{table}
To this end, extensive efforts have been dedicated to social media bias detection \cite{yu2008classifying,iyyer2014political,liu2022politics}, with the primary objective being the prediction of whether a given input (e.g., an article, a paragraph, or a sentence) exhibits bias or not. However, most of previous research focus on fine-tuning models specific to particular datasets \cite{fan2019plain} and subsequently testing them on corresponding test sets. We argue that such trained models lack adaptability and provide predictions that are essentially black-box, lacking in explainability. In this work, we propose a novel bias detection framework based on a comprehensive \textit{bias indicator} database. The term \textit{bias indicator} in this context refers to a concise, descriptive label or tag designed to represent the presence or nature of media bias. Diverging from the coarse-grained framing concept proposed in previous works \cite{card2015media,card2016analyzing,kim2022close}, which cannot be directly applied to bias prediction, our media bias indicators are fine-grained, offering direct insight into the bias exhibited by a given input.

To provide a clearer distinction between framing and our fine-grained media bias indicators, we present several illustrative examples in \Cref{tab:intro}. It becomes evident that framing, exemplified by ``Economic'' and ``Mental health'', falls short in capturing the detailed scope of bias, whereas our fine-grained indicators, automatically generated by LLMs across various dimensions (e.g., tone and language, sources and citations), offer a more comprehensive reflection of bias tendencies. In the context of predicting bias in new text, the prepared bias indicator database can function as a reservoir of human knowledge and experience, while the specific matched indicator can serve as a memory anchor, aiding in the prediction of bias. 

In contrast to much of prior research, which often relies on fine-tuning methods or the training of specific models tailored to particular datasets, leading to limited adaptability and potential performance issues when confronted with out-of-domain data, our IndiVec framework displays notable versatility in bias detection across a wide spectrum of previously unencountered datasets sourced from various origins. Our approach begins with the construction of a bias indicator set, followed by the construction of a vector database based on LLM API. Leveraging the created bias vector database, when processing new text inputs that may contain bias, our bias prediction framework initially extracts or summarizes descriptors based on the given input. Subsequently, these descriptors are matched with indicators stored in the database. The bias label associated with the top-matched indicators dictates the final bias label assigned to the input in question. We conduct explorations on various political leaning prediction datasets with different bias levels (i.e., sentence- and article levels), initially constructing our indicator database based on a single dataset (i.e., FlipBias \cite{chen-etal-2018-learning}). The findings demonstrate that our IndiVec method significantly outperforms the ChatGPT baseline on four distinct political leaning datasets (i.e., FlipBias \cite{chen-etal-2018-learning}, BASIL \cite{fan2019plain}, BABE \cite{spinde2022neural}, MFC \cite{card2015media}) with different sources.

Furthermore, our IndiVec framework shows superiority in explainability. When tasked with detecting bias in a new article or sentence, our framework matches the top-k indicators from the indicator database to represent the bias inclination within the given input based on the distance with bias descriptors if given input. The majority label among these top-k indicators is subsequently employed to classify the input. Importantly, these top-k matched indicators can be interpreted as explanations for the bias prediction. They can also function as a valuable tool for aiding humans in annotating bias data, showing the high degree of explainability of our framework.

In brief, the main contributions of this paper are:
\begin{itemize}[leftmargin=*,topsep=2pt,itemsep=2pt,parsep=0pt]

\item We propose a novel bias prediction framework, called IndiVec, which is based on fine-grained media bias indicators and a matching and voting process that departs from conventional classification-based methods. 

\item We construct a bias indicator dataset consisting of over 20,000 indicators, which can serve as a comprehensive resource for predicting media bias in a more adaptable and explainable manner.

\item Further experiments and analysis validate the effectiveness, adaptability, and explainability of our IndiVec framework.

\end{itemize}

\section{Related Work}
\paragraph{Media Bias.}
Media bias is frequently defined as the presentation of information ``in a prejudiced manner or with a slanted viewpoint'' \cite{golbeck2017large}. However, researchers have explored media bias using diverse definitions and within various contexts, including political \cite{liu2022politics}, linguistic bias \cite{spinde2022neural}, text-level context bias \cite{farber2020multidimensional}, gender bias \cite{grosz2020automatic}, racial bias \cite{barikeri2021redditbias}, etc. Though the bias definition and focus vary, the methodologies are generally based on a classification setting. From classical methods (e.g., Naive Bayes, SVM) \cite{evans2007recounting, yu2008classifying, sapiro2019examining} to deep learning models (e.g., RNN) \cite{iyyer2014political} and pretrained language model-based methods (e.g., BERT and RoBERTa) \cite{liu2022politics, fan2019plain}, they are adopted to predict defined labels in a classification manner. In our work, we treat bias classification as a matching process with fine-grained indicators from a constructed database, and the labels of the matched indicators determine the bias label. Our approach represents a departure from conventional classification methodologies and offers a novel perspective on predicting bias in media.

\paragraph{Political Bias.} It refers to a text’s political leaning or ideology, potentially influencing the reader’s political opinion and, ultimately, their voting behavior \cite{huddy2023oxford}. Political Bias detection has been done at different granularity levels: single sentence \cite{chen-etal-2018-learning, card2015media} and article \cite{fan2019plain, spinde2022neural} level. In this work, we conduct experiments on both sentence- and article-level political bias datasets.

\paragraph{Framing.} Framing refers to emphasizing desired aspects of an issue to promote and amplify a particular perspective \cite{entman1993framing}. Framing in news media and social networks has been studied to analyze political polarization \cite{johnson2016all,tsur2015frame,tourni2021detecting}. \citet{kim2022close} propose a multi-task learning model that jointly learns to embed sentence framing language and predict political bias. However, the frames studied in \citet{kim2022close} are still limited and in the form of topic, which lacks of fine-grained semantics and could not be adopted directly to predict bias label. And the multi-task joint learning's promotion is limited and lack adaptability compared to our IndiVec framework.

\paragraph{Recommendation.} Although the bias detection task is typically considered a classification task, our IndiVec solution aims to address bias detection from the perspective of a recommendation task. For instance, in the quotation recommendation task \cite{wang2021continuity,wang2021quotation,wang2022learning,wang2023quotation}, it is common and fundamental to match quotation candidates with the current query based on the learned representations of both candidates and the query. In this context, IndiVec endeavors to solve a classification task using a recommendation-oriented approach.

\section{Methodology}
\begin{figure}[t]
\centering
\includegraphics[width=1\linewidth]{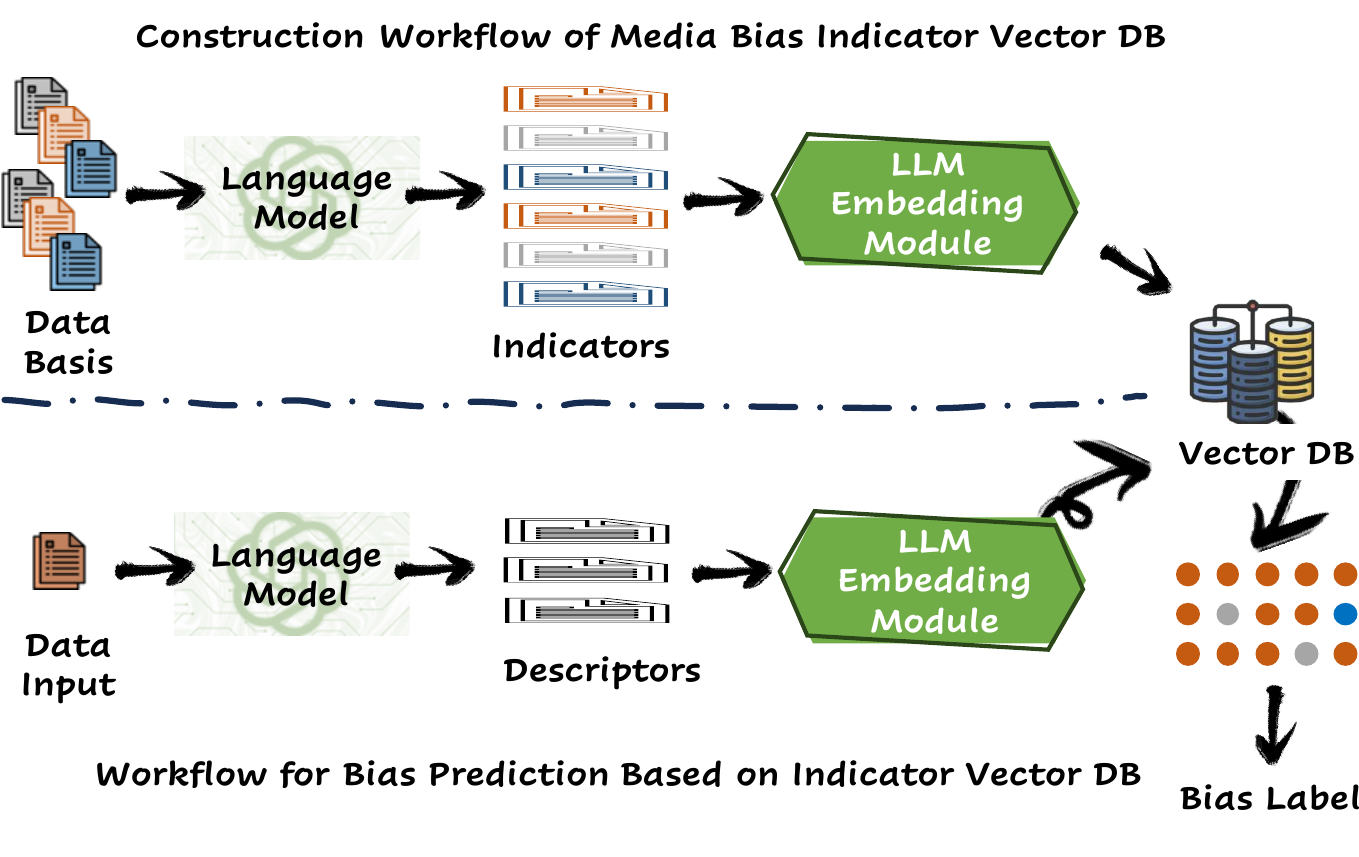}
% \vskip -0.5em
\caption{\label{fig:framework} Our IndiVec Bias Prediction Framework.  }
% \vskip -0.5em
\end{figure}
In this section, we first present the construction of the media bias indicator dataset in \Cref{ssec:model:bias_construction}. Then, we discuss the challenges associated with indicator-based bias prediction and introduce our method of adopting indicators for bias prediction in \Cref{ssec:model:indicator_based_prediction}.

\subsection{Fine-grained Bias Indicator Construction}
\label{ssec:model:bias_construction}
Large Language Models (LLMs) have demonstrated remarkable generative capabilities across various applications and tasks, leveraging their impressive instruction-following capability \cite{qin2023chatgpt}. In this study, we leverage these capabilities by designing meticulously tailored prompts. These prompts will serve as guides for LLMs in the systematic generation of fine-grained labels that accurately reflect the presence of media bias within given articles, text spans, or sentences.
\paragraph{Designing Prompts for Indicator Generation.}
To ensure the precision of indicator generation, we meticulously craft prompts that provide guidance to the LLMs. The objective of prompts is to enable LLMs to assist in analyzing bias or non-bias within input data comprehensively, considering multiple crucial aspects of media bias assessment. The aspects include tone and language, sources and citations, coverage and balance, agenda and framing, and bias in examples and analogies (refer to Table \ref{tab:bias_category}). These aspects collectively contribute to a nuanced understanding of bias within the content. Furthermore, to facilitate LLMs' understanding of these aspects, we incorporate detailed descriptions and illustrative examples into the prompts. Specifically, the prompt is structured as follows:
\begin{quote}
\textit{Demonstration of bias indicator categories: \colorbox{Lightgray}{\color{white}{\textit{DESC\&EX}}}. \\
Based on the demonstration provided above, please label the \colorbox{Lightgray}{\color{white}{\textit{TEXT INPUT}}} with bias indicators to identify the political leaning \colorbox{Lightgray}{\color{white}{\textit{GIVEN LABEL}}}.
}
\end{quote}
\noindent where \colorbox{Lightgray}{\color{white}{\textit{DESC\&EX}}} represents description and examples of indicator categories shown in Table \ref{tab:bias_category}.

\paragraph{Bias Indicator Generation.} When LLMs are guided with the specific prompts we have introduced earlier, they possess the strong instruction-following capability to generate bias indicators. We collect the generated indicators, denoted as $\mathcal{I}_0$, which serve as fundamental components in the further bias assessment process. These indicators enable us to systematically evaluate and categorize media bias, thereby contributing to a more nuanced understanding of bias within the analyzed content.

\paragraph{Verification of Generated Indicator.} 
To ensure the quality of the generated indicators, we adopt a multi-strategy based verification. The strategies include: (1) We eliminate indicators that conflict with the provided ground truth labels. (2) Utilizing Large Language Models (LLM), we conduct a backward verification process and exclude indicators with low confidence in their ability to signify bias or non-bias. 
After verification, we get the indicator set $\mathcal{I} = \{i_1, i_2, ..., i_{|\mathcal{I}|}\}$, and the corresponding bias label for $\mathcal{I}$ is $\mathcal{Y}=\{y_1, y_2, ..., y_{|\mathcal{I}|}\}\; (y_j \in \{0, 1, 2\}, \forall j \in \{1, 2, ..., |\mathcal{I}|\})$.

\subsection{Indicator Enhanced Bias Prediction}
\label{ssec:model:indicator_based_prediction}
Our automatically generated and verified fine-grained indicator set serves as a valuable resource for facilitating the analysis and prediction of bias. In this subsection, we first discuss the potential challenges associated with applying media bias indicators in bias detection. Then, we elaborate on our approach to utilizing the media bias indicator set $\mathcal{I}$ as a foundation for media bias detection.

\paragraph{Challenges in Indicator-based Bias Prediction}
One intuitive approach is to match the input text to the fine-grained indicators, where the bias label for the given input could be the bias label associated with the matched indicators. However, the size of the indicator set is quite large, and this poses a challenge for multi-label classification-based methods due to the sparse output space. Additionally, the semantic space of the indicators differs from that of the normal input text (e.g., input articles or sentence spans to detect bias) since the indicators are concise sentences that are associated with bias labels. Moreover, traditional approaches, such as training from scratch or fine-tuning the indicator matching method \cite{liu2019detecting}, may lead to a lack of adaptability, which deviates from our original goal of enhancing the adaptability of bias prediction.

To address the challenges mentioned above, we propose the utilization of a vector database technique that has recently garnered significant attention among researchers \cite{peng2023embedding}. Initially, we create a vector database based on the indicator set and an off-the-shelf LLM text embedding API. Additionally, we extract descriptors from the input text based on similar prompt using in constructing indicator set (the difference is that we do not provide ground truth bias label), which can be considered as labels or tags within a similar semantic space as the indicators. Finally, we employ a matching process between the descriptors of the input text and the indicators based on their embedding representations' distances. Notably, this approach circumvents the need for additional training efforts and capitalizes on the robust representation extraction capabilities of LLMs. The formal description of our indicator-based bias prediction process is as follows. 

\paragraph{Bias Prediction with Vector Database.}
Based on the maintained indicator set $\mathcal{I}$, we first construct and store the corresponding vector database $\mathcal{V}_{\mathcal{I}}=\{v_1, v_2, ..., v_{|\mathcal{I}|}\}$. Here $v_j \; (j \in \{1,2,...,|\mathcal{I}|\})$ is an N-dimensional vector representing its semantic information derived from techniques of embedding extraction (e.g., OpenAI Embeddings\footnote{\url{https://platform.openai.com/docs/guides/embeddings}}).

\begin{equation}
    v_j \leftarrow Embed(i_j), j \in \{1, 2, ..., |\mathcal{I}|\} 
\end{equation}

Given one query text input noted as $c$, we first generate its descriptor $\mathcal{D}^{c}=\{d_1^c, d_2^c, ..., d_{|\mathcal{D}^c|}^c\}$. For each $d_j^c \in \mathcal{D}^{c}$, we extract its vector representation $v_j^c$ with the identical embedding extraction method. Then, the distance between $v_j^c$ and vectors in the vector database $\mathcal{V}_{\mathcal{I}}$ can be computed using cosine similarity metric:
\begin{equation}
    Distance(v_j^c,v_k) = \frac{v_j^c \cdot v_k}{|v_j^c||v_k|}
\end{equation}
where $k \in \{1,2,...,|\mathcal{I}|\}$. For each descriptor $d_j^c \in \mathcal{D}^c$, we rank the $|\mathcal{I}|$ bias indicators based on their distances to $d_j^c$ and extract the top $M$ bias indicators. Here, $M$ is a hyper-parameter. The corresponding bias labels for these selected $M$ bias indicators are denoted as $\{y_{j,1}^c, y_{j,2}^c, \ldots, y_{j,M}^c\}$. Finally, we predict the bias label for input $c$ using majority voting. In other words, the bias label assigned to query $c$ is determined by the majority value among the $|\mathcal{D}^c| \times M$ labels.

\section{Experimental Setup}
\label{sec:exp_setup}
\paragraph{Datasets.}
\begin{table*}[t]
\setlength{\tabcolsep}{1mm}\small
\begin{center}
\resizebox{\linewidth}{!}{
\begin{tabular}{p{2.5cm}|c|p{3.5cm}|p{2.5cm}|c|c|c|c}
\toprule[1.0pt]

Dataset & Bias Level & Source & Bias Label & Paired & \# of Instances & Avg Length & \% of Biased Instances\\

\midrule[0.5pt]
FlipBias \cite{chen-etal-2018-learning}& Article & New York Times, Huffington Post, Fox News and Townhall & Left, Center, Right & Yes & 6,447 & 909 & 76.5 \% \\
\midrule[0.5pt]
BASIL \cite{fan2019plain} & Sentence & Huffington Post, Fox News, and New York Times & Lexical Bias, Informational Bias & Yes & 7,984 & 24.1 & 19.6\% \\
\midrule[0.5pt]
BABE \cite{spinde2022neural} & Sentence & Fox News, Breitbart, Alternet and so on & Biased, Non-biased & No & 3,674 & 32.6 & 49.3\% \\
\midrule[0.5pt]
MFC (V2) \cite{card2015media} & Article & Lexis-Nexis (Database) & Pro, Neutral, Anti & No & 37,623 & 260 & 84.5\% \\
\bottomrule[1.0pt]
\end{tabular}
}
\end{center}
\vskip -1em
\caption{\label{tab:dataset} Statistics of the Datasets Used in Experiments: FlipBias, BASIL, BABE, and MFC.
}
\vskip -1em
\end{table*}
Though our media bias prediction framework is applicable for various types of bias, we primarily conducted experiments on political bias datasets due to their higher visibility and greater abundance. In our main experiments, we established a bias indicator vector database based on the FlipBias dataset \cite{chen-etal-2018-learning}.
This dataset was sourced from the news aggregation platform allsides.com in 2018, comprising a total of 2,781 events and each event is represented with sufficient text from different political leanings, providing diverse information and opinions. The data's high quality and wide recognition make it the optimal choice to construct the vector database.
Employing this constructed bias indicator database, in addition to the FlipBias dataset, we evaluated the model's performance on three additional datasets: BASIL \cite{fan2019plain}, BABE \cite{spinde2022neural}, and the Media Frame Corpus (MFC) \cite{card2015media}. We relabeled these datasets as Biased and Non-Biased instances following \citet{wessel2023introducing}. A detailed statistical analysis of these four datasets is provided in \Cref{tab:dataset}. Further elaboration along with examples related to the datasets can be found in \Cref{ssec:app:dataset}.

\paragraph{Comparison Setting.} We compare our IndiVec framework against two types of baselines: \textsc{Finetune}, which involves fine-tuning a pretrained language model \cite{fan2019plain}, and \textsc{ChatGpt}. For the \textsc{Finetune} model, we take into consideration that our bias indicator is constructed exclusively from the FlipBias dataset. To ensure a fair comparison, we fine-tune pretrained language models, specifically BERT \cite{devlin2018bert} and GPT3.5\footnote{https://openai.com/blog/gpt-3-5-turbo-fine-tuning-and-api-updates}, using the training set of the FlipBias dataset. Subsequently, we present the test performance results on the test sets of the four datasets. As for the \textsc{ChatGpt} baseline, we employ zero-shot and few-shot approaches to predict bias labels, where the input data are directly presented with proper prompts to query ChatGPT for bias label prediction.

\paragraph{Evaluation Metrics.} 
In our evaluation, we account for the varying proportions of biased and non-biased instances in the four datasets, which often result in severe label imbalances as shown in \Cref{tab:dataset}. Our assessment of model performance encompasses two key aspects: \textbf{1)} \textit{Precision, Recall, and F1 Score for Biased Instances}: This set of metrics helps us gauge the models' ability to detect bias in the dataset. \textbf{2)} \textit{MicroF1 and MacroF1 for Both Biased and Non-Biased Instances}: These metrics provide insights into the overall prediction capabilities of the models, considering both biased and non-biased instances.

\paragraph{Implementation Details}
When conducting the fine-tuning experiments, we fine-tune the model using the pre-trained BERT model \cite{devlin2018bert} and the AdamW optimizer \cite{loshchilov2017decoupled}. This fine-tuning process was facilitated through Hugging Face \cite{wolf2020transformers}, and we specifically employed the \textit{BertForSequenceClassification} model.

In the implementation related to the large language model, we utilized the \textit{gpt-3.5-turbo-16k} model via LangChain. The bias indicators are transformed into vectors using the text embedding model \textit{text-embedding-ada-002}. These vectors are stored in the Chroma vector database, which is hosted on our local machine. The database acts as the search library for identifying similar vectors in the indicator matching process.

In the process of indicator verification, we prompt \textit{gpt-3.5-turbo-16k} model for the confidence score (a number from 1 to 10) of each indicator. The average confidence score of our 24,272 indicators is 6.82. Consequently, we obtained 19,377 indicators after filtering the indicators with confidence scores less than 6. 
When predicting bias with vector database, our hyper-parameter $M$ is set to $10$, and the average numbers of descriptors $|\mathcal{D}^c|$ are $4.0, 2.7, 3.3, 4.2$ in FlipBias, BASIL, BABE, and MFC.
Besides, we also conduct Left-Center-Right 3-way classification on dataset Flipbias and ABP \cite{baly2020we}. 

\section{Experimental Results}
\subsection{Main Comparison Results}

\begin{table*}[t]
\setlength{\tabcolsep}{1.1mm}\small
\newcommand{\tabincell}[2]{\begin{tabular}{@{}#1@{}}#2\end{tabular}}
\begin{center}
\resizebox{\linewidth}{!}{
\begin{tabular}{l|ccccc|ccccc|ccccc|ccccc}
\toprule
\multirow{2}{*}{Base Models} & \multicolumn{5}{c}{ \tabincell{c}{FlipBias}} & \multicolumn{5}{c}{ \tabincell{c}{BASIL}} & \multicolumn{5}{c}{ \tabincell{c}{BABE}} & \multicolumn{5}{c}{ \tabincell{c}{MFC}}
\\
\cmidrule(lr){2-6}\cmidrule(lr){7-11}\cmidrule(lr){12-16}\cmidrule(lr){17-21}
& FT-B & FT-G & G-ZS & G-FS & IndiVec & FT-B & FT-G & G-ZS & G-FS & IndiVec & FT-B & FT-G & G-ZS & G-FS & IndiVec & FT-B & FT-G & G-ZS & G-FS & IndiVec \\
\midrule
\rowcolor[rgb]{0.93,0.93,0.93} 
    \multicolumn{21}{c}{\textit{Scores on Biased Instances}} \\
Precision & 83.6 & \textbf{88.7} & 63.9& 59.9 & 62.7 & 19.1 & 20.0 & \textbf{39.3} & 22.4 & 32.2 & 49.2 & 37.7 & \textbf{81.9} & 53.7 & 62.9 & 86.3 & 85.8 &86.5 & 86.4 & \textbf{86.9}\\
Recall &\textbf{98.6} &93.6&22.1&61.4&71.6 & \textbf{100}& 94.9 &2.3 &44.7& 34.9& 99.8 & \textbf{100}&20.1&68.6&78.9 &76.4 &\textbf{95.3}&37.2&72.9& 78.6\\
F1 & 90.5 & \textbf{91.1} &32.9&60.6&66.9&32.0&33.0 &4.4 &29.5&\textbf{33.5}&65.9& 54.7& 32.2&60.2&\textbf{70.0}&81.1 & \textbf{90.3}&52.3 &79.1&  82.5\\
\midrule
\rowcolor[rgb]{0.93,0.93,0.93} 
    \multicolumn{21}{c}{\textit{Scores on Both Biased and Non-Biased Instances}} \\
% \underline{Scores on All Categories} &&&& &&&& &&&& \\
Micro F1 &87.5 &\textbf{90.0}&45.8&52.1&57.2&16.1 &25.0&\textbf{80.7}&59.7&73.7&49.2 & 38.0&58.4&55.4&\textbf{66.7} &69.3&\textbf{82.5}&41.0&66.8& 71.4 \\
Macro F1  &\textbf{89.9}&89.8 &43.7&49.8&53.2&19.1 &23.9&46.8&50.5&\textbf{58.6}&33.0&28.2&51.1&54.7&\textbf{66.3} &50.0&50.3&37.3&49.4& \textbf{51.8}\\

\bottomrule
\end{tabular}
}
\end{center}
\vskip -1em
\caption{\label{tab:main_results} 
Comparison results (in \%) on four datasets. ``FT'' means fine-tuning the bias prediction model using the Flipbias training set, followed by reporting the prediction results on the test sets of the four datasets. ``G'' means the model GPT-3.5, ``B'' means the model BERT, ``G-ZS'' and ``G-FZ'' mean zero-shot and few-shot setting on ChatGPT.
}
\vskip -1em
\end{table*}

\begin{table*}[t]
\setlength{\tabcolsep}{1.1mm}\small
\newcommand{\tabincell}[2]{\begin{tabular}{@{}#1@{}}#2\end{tabular}}
\begin{center}
% \resizebox{\linewidth}{!}{
\begin{tabular}{l|ccc|ccc|ccc|ccc}
\toprule
\multirow{2}{*}{Base Models} & \multicolumn{3}{c}{ \tabincell{c}{FlipBias}} & \multicolumn{3}{c}{ \tabincell{c}{BASIL}} & \multicolumn{3}{c}{ \tabincell{c}{BABE}} & \multicolumn{3}{c}{ \tabincell{c}{MFC}}
\\
\cmidrule(lr){2-4}\cmidrule(lr){5-7}\cmidrule(lr){8-10}\cmidrule(lr){11-13}
& Pre & Rec & F1 & Pre & Rec & F1& Pre & Rec & F1& Pre & Rec & F1\\
\midrule
Full model &62.7&71.6&66.9 &\textbf{32.2}&34.9&\textbf{33.5} & \textbf{62.9}&78.9&\textbf{70.0} &86.9&\textbf{78.6}&\textbf{82.5}\\ 
$\;\;$ - $\mathcal{I}$ construction's Desc\&Ex   &62.9&53.1&57.6&23.9 &52.8&33.0 &57.7&71.7 &63.9&\textbf{87.6}&41.7&56.5\\
$\;\;\;\;\;\;$ - $\mathcal{I}$ construction's verification    &\textbf{64.3}&53.8&58.6 &23.7 &\textbf{59.6}&33.9 &56.0&75.4 &64.3&\textbf{87.6}&46.8&61.0\\
$\;\;\;\;\;\;\;\;\;\;$ - Descriptor mapping    &60.5&\textbf{95.5}&\textbf{74.1} &20.9 &52.3&29.9&49.8&49.8 &49.8& 85.0& 42.5&56.6\\
$\;\;$ - $\mathcal{I}$ construction's verification    &62.2&70.5&66.1&31.9 &29.7 &30.8 &60.4&79.3 &68.5&\textbf{87.6}&68.8&77.1\\
$\;\;\;\;\;\;$ - Descriptor mapping    &61.6&68.1&64.7&28.6 &37.7&32.5&56.9&\textbf{79.5} &66.3&85.9&73.3&79.1\\
% $\;\;$ - $\mathcal{F}_{indicator}$ mapping    &&&& &&&& &&&&\\
\bottomrule
\end{tabular}
% }
\end{center}
\vskip -1em
\caption{\label{tab:ablation_study} 
Ablation study results (in \%) on four datasets. 
}
\vskip -1em
\end{table*}

\begin{table*}[t]
\setlength{\tabcolsep}{1.1mm}\small
\newcommand{\tabincell}[2]{\begin{tabular}{@{}#1@{}}#2\end{tabular}}
\begin{center}
% \resizebox{\linewidth}{!}{
\begin{tabular}{l|ccc|ccc|ccc|ccc}
\toprule
\multirow{2}{*}{Training Set} & \multicolumn{3}{c}{ \tabincell{c}{FlipBias}} & \multicolumn{3}{c}{ \tabincell{c}{BASIL}} & \multicolumn{3}{c}{ \tabincell{c}{BABE}} & \multicolumn{3}{c}{ \tabincell{c}{MFC}}
\\
\cmidrule(lr){2-4}\cmidrule(lr){5-7}\cmidrule(lr){8-10}\cmidrule(lr){11-13}
& F1 & MicroF1 & MacroF1 & F1 & MicroF1 & MacroF1& F1 & MicroF1 & MacroF1& F1 & MicroF1 & MacroF1\\
\midrule
% SOTA &&&& &&&& &&&&\\
% \midrule
\underline{F}lipBias & \cellcolor[rgb]{0.93,0.93,0.93}{\textbf{90.5}} & \cellcolor[rgb]{0.93,0.93,0.93}{\textbf{87.5}} & \cellcolor[rgb]{0.93,0.93,0.93}{\textbf{86.2}} & 32.0 & 16.1 & 19.1 & 65.9 & 49.2 & 33.0 & 81.1 & 69.3 & 50.0 \\
\underline{B}ASIL & 1.6 & 40.4 & 29.4 & \cellcolor[rgb]{0.93,0.93,0.93}{\textbf{48.4}} & \cellcolor[rgb]{0.93,0.93,0.93}{\textbf{83.3}} & \cellcolor[rgb]{0.93,0.93,0.93}{69.2} & 57.1 & 66.4 & 64.7 & 2.6 & 15.0 & 13.6 \\
\underline{B}ABE & 41.2 & 48.3 & 39.7 & 31.1 & 69.7 & 55.7 & \cellcolor[rgb]{0.93,0.93,0.93}{\textbf{72.7}} & \cellcolor[rgb]{0.93,0.93,0.93}{\textbf{75.2}} & \cellcolor[rgb]{0.93,0.93,0.93}{\textbf{74.9}} & 64.8 & 55.1 & 41.7 \\
\underline{M}FC & 74.8 & 59.8 & 37.6 & 32.1 & 19.0 & 21.1 & 65.9 & 49.5 & 34.2 & \cellcolor[rgb]{0.93,0.93,0.93}{\textbf{92.6}} & \cellcolor[rgb]{0.93,0.93,0.93}{\textbf{86.5}} & \cellcolor[rgb]{0.93,0.93,0.93}{56.8} \\
All (\underline{FBBM})  & 89.7 & 87.0 & 85.9 & 30.6 & 60.6 & \textbf{83.4} & 70.5 & 74.5 & 74.0 & 91.9 & 85.4 & \textbf{59.4} \\

\bottomrule
\end{tabular}
% }
\end{center}
\vskip -1em
\caption{\label{tab:finetuning_comparison} 
Comparison results (in \%) of models with different finetuning training sets. When we refer to ``BASIL-FlipBias'', it indicates training the model using the BASIL training set and then evaluating on FlipBias test set.
}
\vskip -1em
\end{table*}
We report the main comparison results on four datasets in \Cref{tab:main_results}. We have the following observations based on the main results.

$\bullet$~\textit{Our \textsc{IndiVec} framework demonstrates greater adaptability compared to the \textsc{Finetune} model trained on specific data.} As we introduced in \Cref{sec:exp_setup}, our \textsc{IndiVec} is constructed based on the FlipBias dataset, while \textsc{Finetune} is fine-tuned on the same dataset. Although \textsc{Finetune} exhibits better performance on the in-domain test set (i.e., the FlipBias test set), it shows poorer performance on out-of-domain data (i.e., the test sets of BASIL, BABE, and MFC), particularly on datasets with different data formats (e.g., FlipBias exhibits article-level bias, whereas BASIL and BABE feature sentence-level bias). 
Although the GPT Finetune model outperforms the BERT Finetune model on the in-domain FlipBias test set, together with the same granularity, article-level, dataset MFC. It still cannot work well in imbalanced and out-of-domain data, which shows that the lack of generability is a common shortcoming of finetuning-based methods.
In contrast, our \textsc{IndiVec} demonstrates promising performance for both in-domain and out-of-domain data. To further validate the claim that \textsc{Finetune} cannot handle out-of-domain data effectively, we conducted a more comprehensive set of experiments by fine-tuning the base BERT model \cite{fan2019plain} on four separate datasets, as well as on the combined dataset (referred to as FBMM). The results are presented in \Cref{tab:finetuning_comparison}. From the results, it is evident that even fine-tuning on the combined dataset did not yield the best performance. This further underscores the superiority of our general \textsc{IndiVec} bias detection framework.

$\bullet$~\textit{Our \textsc{IndiVec} framework surpasses \textsc{ChatGpt}}. In addition to its advancements over traditional fine-tuning methods, as shown in \Cref{tab:main_results}, \textsc{IndiVec} consistently outperforms \textsc{ChatGpt} across various evaluation metrics and datasets whether on zero-shot or few-shot setting. These improvements can be attributed to the fine-grained bias vector database, which offers denser knowledge on media bias compared to general large language models such as ChatGPT. 

$\bullet$~\textit{Imbalanced data does not have 
a significant affect on our \textsc{IndiVec} framework.} By observing the microF1 and macroF1 scores on both biased and non-biased instances in \Cref{tab:main_results} and the proportions of biased and non-biased instances listed in \Cref{tab:dataset}, we can find that our \textsc{IndiVec} framework effectively handles datasets, irrespective of the degree of imbalance. This ability may be attributed to the fact that \textsc{IndiVec}'s bias prediction does not rely on training with the target data. 
% We explore the potential factors influencing \textsc{IndiVec}'s performance in \Cref{ssec:exp:effectivenees_biasDB}.
\paragraph{Ablation Study.} To further analyze the effectiveness of the proposed mechanisms, including multi-dimensional considerations in indicator construction, post-verification to enhance the indicator set's quality, and the alignment of semantic space between normal sentences and indicators through mapping, we conducted an ablation study and present the results in \Cref{tab:ablation_study}. We find:

$\bullet$~\textit{All proposed mechanisms are effective especially on out-of-domain data.} By examining the ablation results of the variations to our full model in \Cref{tab:ablation_study}, it becomes evident that all the proposed mechanisms have a positive impact on performance in out-of-domain data (BASIL, BABE, MFC). 

When analyzing the results on FlipBias, we observe that the highest F1 achieved by the simplest variant is attributed to an extremely high Recall score (e.g., 95.5 Recall, indicating a preference for labeling most test data as biased). 
It indicates that our components help to construct more general indicators instead of domain-specific indicators, which could generally perform well across all datasets.

$\bullet$~\textit{Both the diversity and quality of indicators play a vital role.} When we analyze the outcomes of our complete model and its variants, which exclude the ``Desc\&Ex'' category during indicator construction (potentially reducing indicator diversity), it becomes evident that the effective presentation of indicators leads to improved prediction performance. This enhancement can be attributed to the fact that a well-crafted presentation can facilitate the generation of higher-quality, more varied indicators from various dimensions, thereby bolstering prediction accuracy. Additionally, when we assess the results of our full model and its variants that exclude backward verification, it becomes apparent that higher-quality indicators can significantly enhance bias prediction performance.

\subsection{Effectiveness of Bias Indicator Vector DB}
\label{ssec:exp:effectivenees_biasDB}
\paragraph{Statistic of the constructed indicators.}
\begin{figure}[t]
\centering
\includegraphics[width=0.65\linewidth]{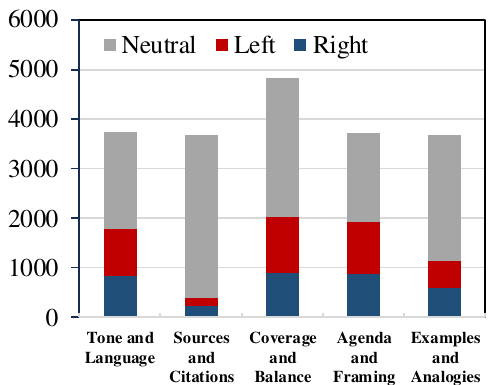}
\vskip -0.5em
\caption{\label{fig:indicator_statistic} Statistics of Constructed Indicator Set.  }
\vskip -0.5em
\end{figure}
Before we explore assessing the effectiveness of our media bias indicator vector database (referred to as IndiVecDB), we first present statistics about the indicators annotated by LLM in \Cref{fig:indicator_statistic}. It's evident that the indicator numbers across different categories are generally well-balanced. However, there are significant differences in the distribution of political leanings among the various categories. Notably, indicators in the ``\textit{Sources and Citations}'' and ``\textit{Examples and Analogies}'' categories tend to exhibit a neutral stance. This suggests that articles or sentences marked with specific sources, citations, and examples are more likely to be neutral. Furthermore, we conducted a statistical analysis of the length of the constructed indicators, revealing an average length of 15.9 tokens per indicator. This length is notably longer than the framing discussed in previous work \cite{fan2019plain}, while also conveying richer semantics.

\definecolor{left-leaning}{RGB}{220,20,60} 
\definecolor{right-leaning}{RGB}{0,0,128}   
\definecolor{neutral}{RGB}{100,100,100}   
\begin{table*}[t]
\setlength{\tabcolsep}{1mm}\small
\begin{center}
\resizebox{\linewidth}{!}{
\begin{tabular}{c|m{6.5cm}|m{4cm}|m{4cm}|m{1cm}}
\toprule[1.0pt]

\textbf{Dataset} & \textbf{Input Text} & \textbf{Generated Descriptor} & \textbf{Top-1 Matched Indicator} & \textbf{Ground Label}\\

\midrule[0.5pt]
\multirow{5}{*}{\textbf{BABE}} & \multirow{2}{\linewidth}{A Joe Biden presidency could reset ties with top U S trade partner Mexico that have suffered since Donald Trump made his first White House bid tarring Mexican migrants as rapists and gun runners and vowing to keep them out with a border wall}
 & Describes Donald Trump's statements negatively& \textcolor{left-leaning}{Uses negative language to describe Donald Trump's actions and behavior }& \multirow{6}{*}{Biased}  \\
 \cmidrule(lr){3-4}
& &Frames Trump's statements as damaging to US-Mexico ties & \textcolor{right-leaning}{Trump's criticism of Mexico, negative language towards trade actions} & \\
\midrule[0.5pt]
\multirow{6}{*}{\textbf{MFC}} &\multirow{2}{\linewidth}{Village calls for stricter gun control State law limits Royal Palm Beach ... for lawmakers to enact stricter gun measures in the wake of ... But they ve lamented that their hands are tied by a 2011 Flor ida law that punishes local governments that try to pass their own gun control rules ... get us into the details that the current version does he said adding that he would prefer something general yet comprehensive} & "stricter gun measures" and "punishes local governments” & \textcolor{left-leaning}{Emotional appeals for stricter gun laws and criticism of politicians who oppose them} & \multirow{6}{*}{Biased}  \\
 \cmidrule(lr){3-4} 
& &No specific sources or citations provided&\textcolor{neutral}{No specific sources or citations provided}& \\
 \cmidrule(lr){3-4} 
& & Presents the council's call for stricter gun control as a response to the Parkland shooting & \textcolor{left-leaning}{Focuses on the need for stronger gun controls and the opposition from the gun lobby} & \\
\midrule[0.5pt]
\textbf{FlipBias} & LAUSANNE, Switzerland (Reuters) - Russia has been banned from the 2018 Pyeongchang Winter Olympics after the IOC found evidence … &
Describes the evidence of "unprecedented systematic manipulation" and "manipulation of doping and the anti & \textcolor{neutral}{Provides details of the alleged robbery and the athletes' actions } & Non-Biased \\
 % \cmidrule(lr){3-4}
% & &Frames Trump's statements as damaging to US-Mexico ties & \textcolor{right-leaning}{Trump's criticism of Mexico, negative language towards trade actions} & \\

\bottomrule[1.0pt]
\end{tabular}
}
\end{center}
\vskip -1em
\caption{\label{tab:case_study} Sentence- and article-level biased examples from BABE, MFC, and FlipBias datasets, with Indicators in \textcolor{neutral}{Gray}, \textcolor{left-leaning}{Red}, and \textcolor{right-leaning}{Blue} representing associated bias labels (\textcolor{neutral}{Gray} for Neutral, \textcolor{left-leaning}{Red} for Left-Leaning, \textcolor{right-leaning}{Blue} for Right-Leaning).
}
% \vskip -1em
\end{table*}
\paragraph{Case Study.} We present case studies involving two examples selected from the BABE, MFC, and FlipBias datasets, as shown in \Cref{tab:case_study}. These case studies highlight the role of the generated descriptors and matched indicators in assessing bias at both article and sentence levels. 
For lengthy sequences, as the example from the MFC dataset in \Cref{tab:case_study}, where humans might not quickly locate bias, our generated descriptors are explainable and visible for end-users, making it particularly crucial for article-level bias detection.

In contrast, their influence on detecting sentence-level bias, as illustrated by the example from the BABE dataset, is less pronounced. These generated descriptors effectively extract and summarize potential bias points from the input, while the matched indicators from our constructed indicator set provide additional insights into bias prediction. Furthermore, upon closer examination of the example from the BABE dataset in \Cref{tab:case_study}, we find that the ground truth bias label for the given input is not always appropriate, as the example does not exhibit obvious bias. In such cases, our \textsc{IndiVec} framework serves as a valuable tool for analyzing potential bias in a more explicit manner. This capability can be especially useful for human annotators when re-evaluating and re-labeling datasets.

\begin{figure}[t]
\centering
\subfigure[Comparison of DB Size] {\label{sfig:DBsize}
\includegraphics[width=0.65\linewidth]{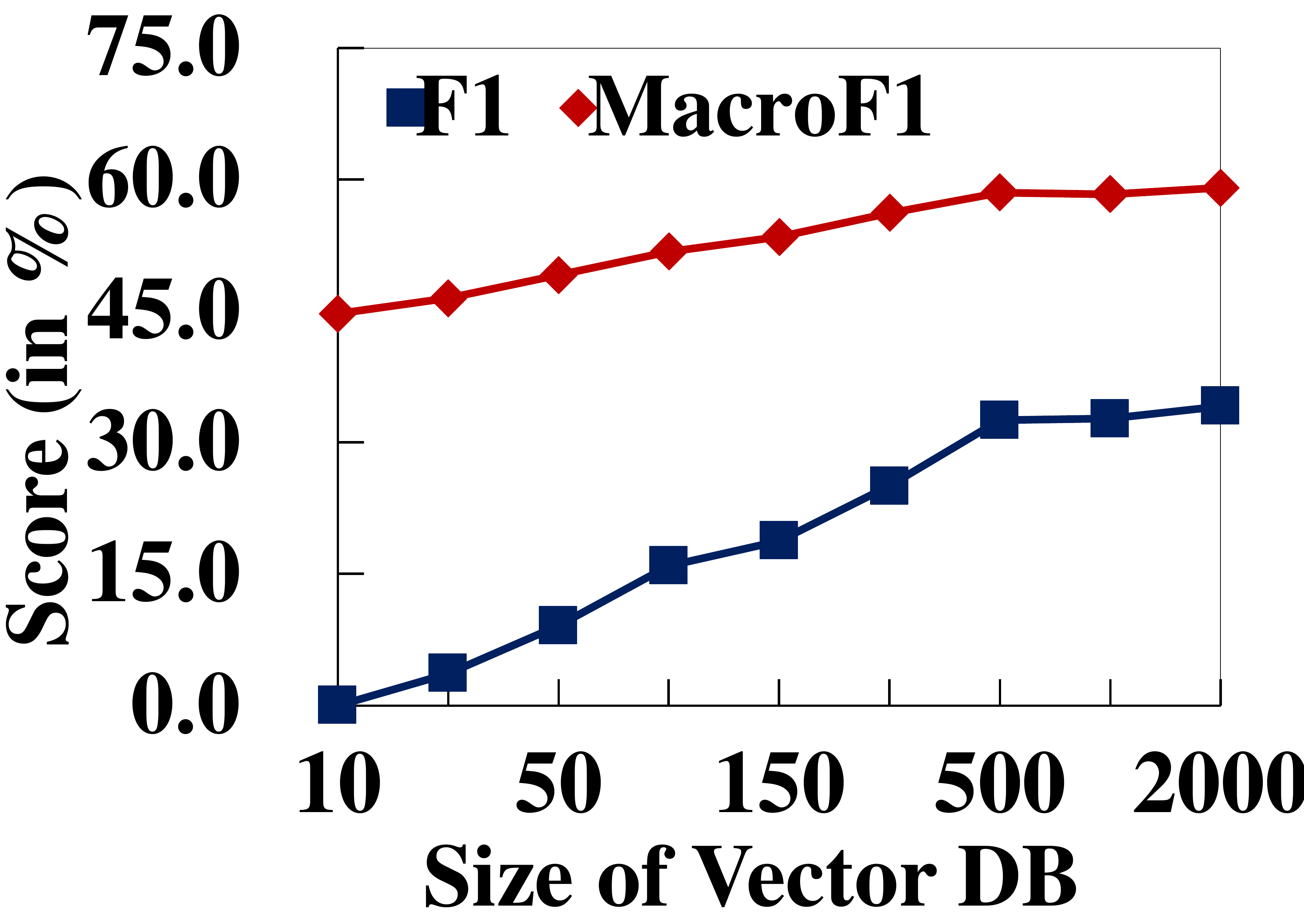}
}
\subfigure[Comparison of Dataset] {\label{sfig:basedataset}
\includegraphics[width=0.65\linewidth]{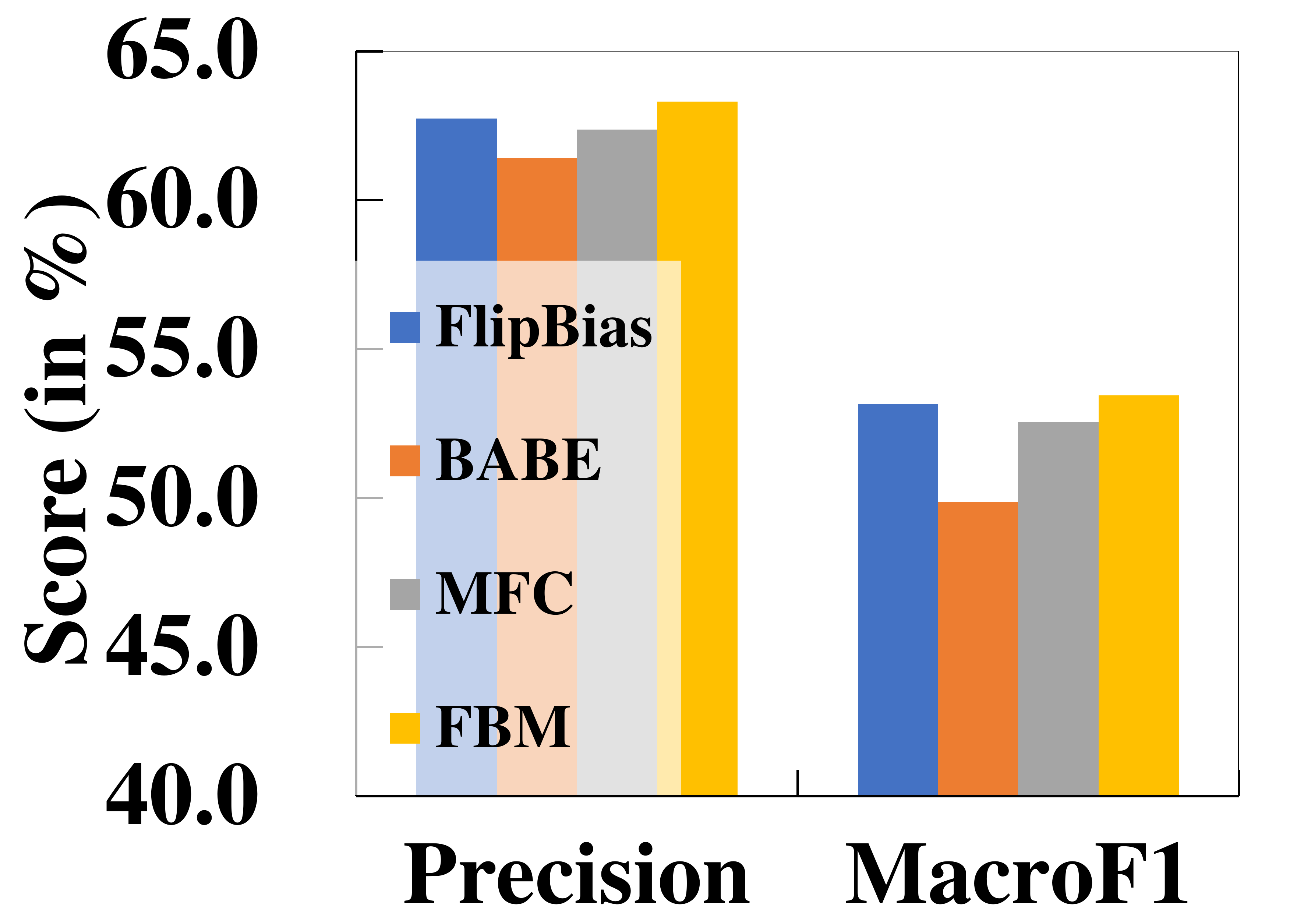}
}
\vskip -1em
\caption{Performance Across Different Indicator Vector Database Sizes (\Cref{sfig:DBsize}) and Varied Base Datasets for Indicator Construction (\Cref{sfig:basedataset}).
}
\vskip -1em
\end{figure}
\paragraph{Effects of Indicator Numbers.} Here we investigate the influence of the number of indicators within the vector database on indicator matching. We systematically vary the number of indicators while maintaining it as a fixed quantity and present the corresponding F1 scores (calculated exclusively for biased instances, as explained in \Cref{tab:main_results}) and MacroF1 scores on the BASIL dataset in \Cref{sfig:DBsize}. Our analysis reveals that as the size of the vector database increases, the overall performance shows a consistent upward trend. Notably, we observe that the performance achieved with a database containing 500 indicators approaches the performance of our full model. This observation suggests that, for a specific test set, there exists a threshold beyond which adding more indicators to the database does not significantly improve performance. However, it is important to note that to accommodate various test sets with different sources, a larger and more diverse database is undoubtedly essential.

\paragraph{Impact of Indicator Diversity.} In our main results (\Cref{tab:main_results}), we rely on indicators constructed from the FlipBias dataset. In this section, we extend our analysis to include indicators derived from various base datasets, including FlipBias, BABE, MFC, and a combination denoted as FBM (comprising FlipBias, BABE, and MFC). We present the precision and MacroF1 results on the FlipBias test set in \Cref{sfig:basedataset}. We can observe that indicators based on the BABE and MFC datasets exhibit relatively lower performance, and the combination FBM does not yield a significant better performance than FlipBias. This may due to that FlipBias is already a diverse and comprehensive data base, and BABE and MFC do not provide additional indicators to help predict bias lables. Intriguingly, even when using a relatively small base dataset like BABE, which comprises only 3674 instances, the MacroF1 score on the test set surpasses that of ChatGPT (as referenced in the results in \Cref{tab:main_results}).

\subsection{Further Analysis}
\label{ssec:exp:further_analysis}
\begin{figure}[t]
\centering
\subfigure[50 Instances (S, D, I)] {\label{sfig:mapping}
\includegraphics[width=0.65\linewidth]{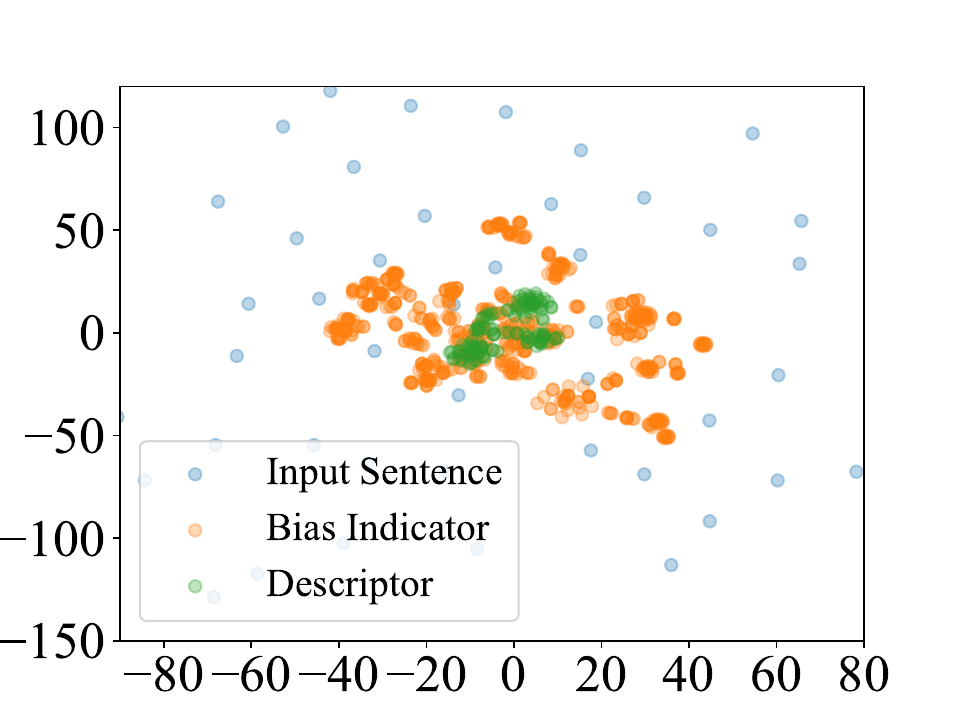}
}
\subfigure[Top and Last Indicators] {\label{sfig:toplast}
\includegraphics[width=0.65\linewidth]{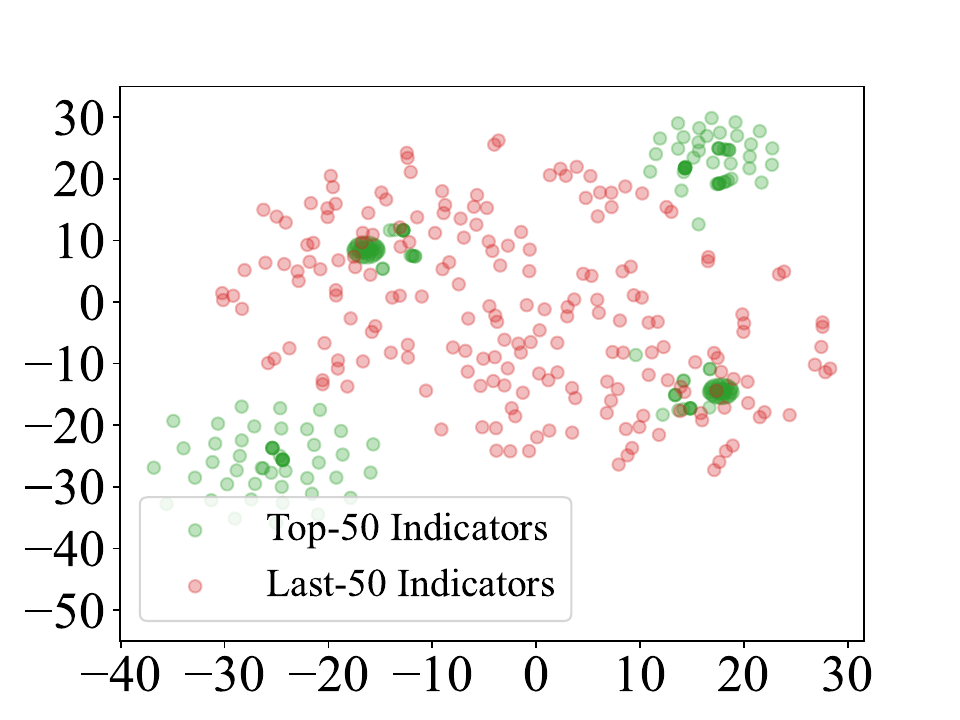}
}
\vskip -1em
\caption{\Cref{sfig:mapping}:Visualization of 50 randomly sampled instances (\underline{S}entence, corresponding \underline{D}escriptor and Top 5 ranked \underline{I}ndicators). \Cref{sfig:toplast}: Visualization of top 50 and last 50 ranked indicators for a randomly selected instance with four Descriptors.
}
\vskip -1em
\end{figure}
In this subsection, we adopt t-SNE \cite{wattenberg2016use} tool to reduce the dimensionality of embeddings from 1536 to 2 and then plot the embeddings in 2D scatter plots to further analyse the effectiveness of our framework.

\paragraph{Difference Between Regular Sentences, Descriptors, and Indicators.} To explore the distinction between regular sentences, descriptors, and indicators, we randomly select 50 sentence inputs from the BABE dataset. Subsequently, we created descriptors and their corresponding top-5 matched indicators for these instances. In \Cref{sfig:mapping}, we present a visual representation of these 50 sentence inputs alongside their descriptors and indicators. We can see that the distribution of the sentence inputs appears random, whereas the descriptors and indicators exhibit clear clustering patterns. Moreover, it's evident that the matched indicators typically reside at the center of the descriptors, aligning with our cosine similarity-based matching procedure. The difference between regular sentence inputs and their descriptors and indicators underscores the necessity of mapping normal inputs to descriptors, as descriptors tend to yield easier matches with indicators.

\paragraph{Difference Between Top-Ranked Indicators and Lower-Ranked Indicators.} To investigate the disparity between top-ranked indicators and those with lower rankings, we selected a random test instance from the BABE dataset. Subsequently, we generated descriptors and matched indicators for these descriptors. In \Cref{sfig:toplast}, we illustrate the top 50 matched indicators alongside the last 50 ranked indicators for this specific instance. Notably, the top-ranked indicators form four distinct clusters, each corresponding to one of the four generated descriptors, while the lower-ranked indicators exhibit a more random distribution.

\section{Conclusion}
This work introduces IndiVec, a novel bias prediction framework. IndiVec leverages fine-grained media bias indicators and employs a unique matching and voting process. We also contribute a bias indicator dataset, encompassing over 20,000 indicators. Our comprehensive experiments and analyses further confirm the effectiveness, adaptability, and explainability of the IndiVec framework, highlighting its potential as a valuable tool for bias detection in media content.

\section*{Limitations}
The limitations of this work are primarily twofold. Although our approach demonstrates high adaptability compared to conventional classification-based and fine-tuning methods, IndiVec remains strongly reliant on the quality and diversity of the base dataset used for constructing the indicator database. While we incorporate multi-dimensional considerations for constructing indicators that can accommodate political datasets from various sources, it's worth noting that these indicators remain focused on political bias and stance-related aspects. In future developments, it would be valuable to explore the creation of indicators based on diverse media bias datasets, not limited to political bias.

Additionally, it's important to acknowledge that the bias labels associated with the generated indicators may not always be accurate. This issue can be attributed to two main reasons. Firstly, as we demonstrated in the case study in \Cref{ssec:exp:effectivenees_biasDB}, the ground truth bias labels of instances can be incorrect, which directly impacts the bias label assigned to the generated bias indicators. Secondly, the generative capabilities of large language models do not always ensure a perfect distinction between neutral and biased content, even after our multi-strategy post-verification and filtering. To address this, more comprehensive and intricate methods may be necessary, especially in real-world applications. This could potentially involve the incorporation of human annotators or the utilization of recent reinforcement learning techniques that incorporate AI feedback mechanisms to enhance the accuracy of bias labels associated with indicators.

\section*{Acknowledgements}
This research work was partially supported by CUHK direct grant No. 4055209, CUHK under Project No. 3230377 (Ref. No. KPF23GW). Jing Li is supported by NSFC Young Scientists Fund (62006203). We are also grateful to the anonymous reviewers for their comments.

\bibliography{anthology,custom}
\bibliographystyle{acl_natbib}

% \newpage
\appendix
% \section*{Appendix}
\section{Detailed Experimental Setup}
\label{sec:app:setup}
\subsection{Details of Datasets}
\label{ssec:app:dataset}

In this subsection, we provide additional details about the datasets used in our experiments.

\paragraph{FlipBias} This dataset \cite{chen-etal-2018-learning} was collected from the news aggregation platform allsides.com in 2018 and comprises a total of 2,781 events. Each event is associated with 2-3 articles from different political leanings, including left, center, and right perspectives. We utilized the sets that encompass both left and right biases simultaneously to generate the bias indicators. The remaining 1,228 articles were reserved for testing purposes. Articles with left or right-leaning perspectives were categorized as biased, while those from the center were designated as non-biased.

\paragraph{BASIL}
BASIL, as presented in Fan et al. (2019) \cite{fan2019plain}, comprises 100 sets of articles, with each set containing 3 articles sourced from Huffington Post, Fox News, and New York Times. Lexical bias and informational bias are annotated at the span level. In our evaluation, a sentence is considered biased if it exhibits either lexical bias or informational bias. For our testing, we randomly selected 10\% of this dataset to serve as the test set, and this test set was used in 5 separate evaluations with different random seeds, following the approach outlined in prior research \cite{van2020context}.

\paragraph{BABE}
BABE, as described in \cite{spinde2022neural}, is a dataset comprising 3,673 sentences sourced from the Media Cloud, an open-source media analysis platform. Expert annotators were tasked with determining whether each sentence exhibited bias or not. To ensure robustness in the results, we conducted a 5-fold cross-validation procedure following the methodology established in prior research \cite{spinde2022neural}.

\paragraph{MFC}
In our research, we utilized the second version of the Media Frame Corpus \cite{card2015media}. This corpus contains a total of 37,622 articles, each of which has been condensed to approximately 225 words and labeled according to the overall tone of the article, which is categorized as either ``pro'', ``neutral'', or ``anti''. Articles with a ``pro'' or ``anti'' tone are considered to exhibit bias.

\definecolor{left}{RGB}{220,20,60} 
\definecolor{right}{RGB}{0,0,128}   
\definecolor{neutral}{RGB}{100,100,100} 
\begin{table*}[t]
  \footnotesize
  % \scriptsize
  \setlength{\tabcolsep}{0.5em}
  \centering

  \begin{tabularx}{\linewidth}{clp{0.6\linewidth}}
    \toprule
\multirow{4}{*}{\textbf{Tone and Language}} &Description & Assess the overall tone of the article, including the choice of words and phrases. Look for emotionally charged language, stereotypes, or inflammatory rhetoric. \\
\cmidrule(lr){2-3}
 &\multirow{3}{*}{Examples} &\textit{\textcolor{left}{Left-leaning}}: The article frequently uses words like "exploitation," "inequality" and "corporate greed" to describe economic issues. \\
 & & \textit{\textcolor{right}{Right-leaning}}: The article employs phrases such as "individual liberty," "free-market solutions," and "personal responsibility" to discuss social policies. \\
 & &\textit{\textcolor{neutral}{Neutral}}: The article maintains a balanced tone without resorting to emotionally charged language or bias-inducing terms.\\
\midrule
\multirow{4}{*}{\textbf{Sources and Citations}} &Description & Check the sources and citations within the article. Assess whether they are from a variety of perspectives or if they predominantly support one side of the political spectrum. \\
\cmidrule(lr){2-3}
 &\multirow{3}{*}{Examples} &\textit{\textcolor{left}{Left-leaning}}: The article primarily cites progressive think tanks, Left-leaning news outlets, and left-wing academics to support its arguments. \\
 & & \textit{\textcolor{right}{Right-leaning}}: The majority of sources cited in the article come from conservative publications, Right-leaning experts, and libertarian think tanks. \\
 & &\textit{\textcolor{neutral}{Neutral}}: The article includes a diverse range of sources from different political backgrounds, providing a balanced set of viewpoints.\\
\midrule
\multirow{4}{*}{\textbf{Coverage and Balance}} &Description & Evaluate whether the article provides a balanced view of the topic or if it tends to favor one particular perspective. \\
\cmidrule(lr){2-3}
 &\multirow{3}{*}{Examples} &\textit{\textcolor{left}{Left-leaning}}: The article predominantly highlights the challenges faced by marginalized communities without sufficiently exploring counterarguments or alternative viewpoints. \\
 & & \textit{\textcolor{right}{Right-leaning}}: The article focuses on the benefits of reduced government intervention without adequately addressing potential drawbacks or opposing viewpoints. \\
 & &\textit{\textcolor{neutral}{Neutral}}: The article presents a comprehensive examination of the topic, addressing both supporting and opposing arguments with equal weight.\\
\midrule
\multirow{4}{*}{\textbf{Agenda and Framing}} &Description & Determine if the article promotes a specific political agenda or frames the issue in a way that aligns with a particular ideology. \\
\cmidrule(lr){2-3}
 &\multirow{3}{*}{Examples} &\textit{\textcolor{left}{Left-leaning}}: The article frames climate change as an urgent crisis requiring immediate government intervention and portrays regulation as the solution. \\
 & & \textit{\textcolor{right}{Right-leaning}}: The article frames tax cuts as essential for economic growth and suggests that limited government intervention is the key to prosperity. \\
 & &\textit{\textcolor{neutral}{Neutral}}: The article objectively presents facts and allows readers to draw their own conclusions without pushing a specific agenda.\\
\midrule
\multirow{4}{*}{\textbf{Examples and Analogies}} &Description & Examine if the article uses examples or analogies that may be biased or misleading in their political implications. \\
\cmidrule(lr){2-3}
 &\multirow{3}{*}{Examples} &\textit{\textcolor{left}{Left-leaning}}: The article compares income inequality to a "wealth gap chasm" and uses emotionally charged analogies to convey the severity of the issue.\\
 & & \textit{\textcolor{right}{Right-leaning}}: The article uses the analogy of a "burdened taxpayer" to describe the negative impacts of government spending.\\
 & &\textit{\textcolor{neutral}{Neutral}}: The article avoids using biased or emotionally charged examples or analogies, sticking to objective and relevant comparisons.\\

\bottomrule
\end{tabularx}
  \caption{Summary of Category of Bias to Guide the Generation of Indicators. 
  }
  \label{tab:bias_category}
\end{table*}
\section{Detailed Indicator DB Construction}
In this section, we provide a detailed explanation of the five categories mentioned to guide the generation of multi-dimension considered indicators, as shown in \Cref{tab:bias_category}. For each category, we offer a concise description and provide examples to facilitate a better understanding of the predefined categories for large language models.

\end{document}